\documentclass[a4paper,11pt]{amsart}
\usepackage{amsthm,amsmath}
\usepackage{graphicx}
\usepackage{latexsym,amssymb}
\usepackage{todonotes}
\usepackage{enumerate}
\usepackage{url}

\theoremstyle{plain}
\newtheorem{lem}{Lemma}

\theoremstyle{definition}
\newtheorem{defn}{Definition}

\theoremstyle{remark}
\newtheorem{rem}[lem]{Remark}

\title{Mechanism Singularities revisited from an algebraic viewpoint}
\author{Zijia Li and Andreas M\"uller}

\begin{document}

\begin{abstract} It has become obvious that certain singular phenomena
cannot be explained by a mere investigation of the configuration space,
defined as the solution set of the loop closure equations. For example, it
was observed that a particular 6R linkage, constructed by combination of two
Goldberg 5R linkages, exhibits kinematic singularities at a smooth point in
its configuration space. Such problems are addressed in this paper. To this
end, an algebraic framework is used in which the constraints are formulated
as polynomial equations using Study parameters. The algebraic object of
study is the ideal generated by the constraint equations (the constraint
ideal). 

Using basic tools from commutative algebra and algebraic geometry (primary
decomposition, Hilbert's Nullstellensatz), the special phenomenon is related
to the fact that the constraint ideal is not a radical ideal. With a primary
decomposition of the constraint ideal, the associated prime ideal of one
primary ideal contains strictly into the associated prime ideal of another
primary ideal which also gives the smooth configuration curve. This analysis
is extended to shaky and kinematotropic linkages, for which examples are
presented.
\end{abstract}

\maketitle

\setcounter{tocdepth}{1}
% \tableofcontents

\section{Introduction}

Kinematic singularities of mechanisms are not always reflected in the
differential geometry of the configuration space (c-space) \cite%
{JMR2018,JMR2019}. Furthermore, certain properties such as the shakiness of
a mechanism, have no corresponding feature in the c-space. Yet any kinematic
phenomenon is somehow encoded in the defining constraint equations. Thus,
the algebraic geometry of the ideal defined by them should allow revealing
such phenomena, when they are defined as polynomial equations. The latter is
always possible for algebraic joints, in particular linkages comprising
revolute and prismatic joints. Algebraic formulations for kinematic modeling
and related approaches for the mechanism analysis have been established \cite%
{Husty10,Li15q}. Still, the relation of special kinematic situations and
their consequences for the algebraic variety and its ideal is not fully
understood. This is addressed in the present paper. The central object for
this study is the ideal generated by the loop constraints.

\section{Algebraic Modeling of Kinematics}

\subsection{Notations}

We use the classical concepts and definitions of dual quaternions and the
Study quadric for kinematics computation from \cite%
{Husty10,Hegedues13f,Hegedues13b,Li15phd}. Dual numbers are denoted by $%
\mathbb{D}:=\mathbb{R}+\epsilon \mathbb{R}$, with multiplication defined by $%
\epsilon ^{2}=0$. Quaternions are denoted by $\mathbb{H}:=<1,\mathbf{i},%
\mathbf{j},\mathbf{k}>_{\mathbb{R}}$ where $\mathbf{i}^{2}=\mathbf{j}^{2}=%
\mathbf{k}^{2}=\mathbf{i}\mathbf{j}\mathbf{k}=-1$, and $\mathbb{D}\mathbb{H}%
:=\mathbb{D}\otimes _{\mathbb{R}}\mathbb{H}$. The conjugate of a dual
quaternion $h$ is denoted by $\overline{h}$ which is obtained by multiplying
the vectorial part of $h$ by $-1$.

Following the definition of Study quadric, we can define the projective $%
\mathbb{D}\mathbb{H}$ which is a real 8-dimensional vector space to obtain $%
\mathbb{P}^{7}$. The Study condition is that $h\overline{h}$ is strictly
real, i.e., its dual part is zero, and is a homogeneous quadratic equation.
In vector notation, the 8-dimensional component vector corresponding to a
dual quaternion $a_{0}+a_{1}\mathbf{i}+a_{2}\mathbf{j}+a_{3}\mathbf{k}%
+\epsilon a_{4}+\epsilon a_{5}\mathbf{i}+\epsilon a_{6}\mathbf{j}+\epsilon
a_{7}\mathbf{k}\in \mathbb{D}\mathbb{H}$ is $\left(
a_{0},a_{1},a_{2},a_{3},a_{4},a_{5},a_{6},a_{7}\right) \in {\mathbb{R}}^{8}$%
. 

\subsection{Loop Closure Constraints of Revolute Joint Linkages}

\paragraph{Constraints in terms of rotation axes:}

The geometry of a kinematic chain with only revolute joints is determined by
the rotation axes of the $n$ joints. A nonzero dual quaternion $h$
represents a rotation (around an axis) if and only if $h\overline{h}$ and $h+%
\overline{h}$ are strictly real, and its primal vectorial part is nonzero.
Therefore, for a single loop mechanism with $n$ rotation joints, we can find 
$n$ dual quaternions $h_{1},\ldots ,h_{n}$ for representing the rotation
axes. With the isomorphism described in \cite[Sect.~2.4]{Husty10}, the
rotation about the $x$-axis and angle $q$ corresponds to the dual quaternion 
$(\cos (\frac{q}{2})-\sin (\frac{q}{2})\mathbf{i})$, which is projectively
equivalent to $(1-\tan (\frac{q}{2})\mathbf{i})$. A formulation of the
closure equation of a single loop mechanism in terms of dual quaternions is
given as%
\begin{equation}
F:=(1-t_{1}h_{1})(1-t_{2}h_{2})\cdots (1-t_{n}h_{n})\equiv 1,  \label{eq:c1}
\end{equation}%
where $t_{i}:=\tan (\frac{q_{i}}{2})$, and $h_{1},\dots ,h_{n}$ are dual
quaternions specifying the rotation axes in the initial position of the
robot, $h_{s}^{2}=-1$ for $s=1,\ldots ,n$. The notation \textquotedblleft $%
\equiv $\textquotedblright\ means projectively equivalent.

\paragraph{Constraints in terms of DH-parameters:}

If the geometry of the kinematic loop is specified by Denavit-Hartenberg
parameters (twist angles, orthogonal distances, offsets and rotation angles) 
\cite{Denavit1955}, the closure constraints are expressed as%
\begin{equation}
F:=(1-t_{1}\mathbf{i})g_{1}(1-t_{2}\mathbf{i})g_{2}\cdots (1-t_{n}\mathbf{i}%
)g_{n}\equiv 1,  \label{eq:c2}
\end{equation}%
where 
\begin{equation}
g_{i}=\left( 1-\frac{s_{i}}{2}\epsilon \mathbf{i}\right) \left( 1-w_{i}%
\mathbf{k}\right) \left( 1-\frac{d_{i}}{2}\epsilon \mathbf{k}\right) ,
\label{gi:1}
\end{equation}%
for $i=1,\dots ,n$. The DH-parameter $\phi _{i}$ is defined as the angle
between the direction vectors of the rotation axis joint $i$ and $i+1$,
denoted with $l_{i}$ and $l_{i+1}$, respectively. For later use we introduce
and $w_{i}=\tan (\frac{\phi _{i}}{2})=\frac{\sin (\phi _{i})}{\cos (\phi
_{i})+1}$. The parameter $d_{i}$ is defined as the orthogonal distance of
the lines $l_{i}$ and $l_{i+1}$. Note that $d_{i}$ may be negative, which
depends on the choice of orientation of the common normal, which we denote
by $n_{i}$. Finally, we define the offset $s_{i}$ as the signed distance of
the intersections of the common normals $n_{i-1}$ and $n_{i}$ with $l_{i}$.
Thus, (\ref{eq:c2}) is just the reformulation of the closure equations (\ref%
{eq:c1}) using DH-parameters in terms of dual quaternions.

\begin{rem}
\label{multiloop}For a multiple-loop mechanism, constraints are formualted
for the topologically independent loops (fundamental cycles) \cite%
{Robotica2018}. The constraint ideal is then the sum of the ideals
associated to the fundamental cycles.
\end{rem}

\begin{rem}
\label{lt} Here we only take an inhomogeneous formulation for the revolute
joints. We constraint the configuration over projective spaces to an affine
space by replacing $(s_{i}-t_{i}h_{i})$ with $(1-t_{i}h_{i})$. When doing
so, some special configurations may not be included, e.g., rotations about
180 degrees. Such configurations can be included using the linear
transformation (M\"{o}bius transformation) $t\mapsto \frac{at+b}{ct+d}$, for
arbitrary real numbers $a,b,c,d$ with $ad-bc\neq 0$. Then we have $%
(1-th)\mapsto (ct+d)-(at+b)h$, which can cover such a configuration in the
computed constraint ideal.
\end{rem}

\subsection{Configuration Space Variety and Constraint Ideal}

With the above formulation, the configuration of a mechanism is represented
by the $n$ joint variables, respectively the tangents, which are summarized
in $\mathbf{t}=(t_{1},\dots ,t_{n})\in {\mathbb{R}}^{n}$. An admissible
configuration must satisfy the algebraic loop closure constraints (\ref%
{eq:c1}) respectively (\ref{eq:c2}). The \emph{configuration space (c-space)}%
, according to this formulation is the algebraic set%
\begin{equation*}
V=\{\mathbf{t\in }\mathbb{R}^{n}|F\left( \mathbf{t}\right) \equiv 1\}.
\end{equation*}%
The c-space will be regarded as an algebraic variety, and the subsequent
analysis will be based on the ideal generated by the constraints, called the 
\emph{constraint ideal}.

To this end, both sides of (\ref{eq:c1}) are written as 8-dimensional
vectors, which yields a system of 7 equations. The constraints are then
given by the components $2,\ldots ,8$, which have to be zero. This gives 7
polynomial equations in the variables $\mathbf{t}=(t_{1},\dots ,t_{n})$. The
5th equation is redundant because it is already satisfied when the other six
equations are fulfilled (due to the Study condition). In order to exclude
'unwanted'\ solutions, i.e. such that $(t_{1}^{2}+1)(t_{2}^{2}+1)\ldots
(t_{n}^{2}+1)\neq 0$. In order to express this condition as an equality
constraint, an extra variable $u$ is introduced and (\ref{eq:c1}) is
complemented by the equation $(t_{1}^{2}+1)(t_{2}^{2}+1)\cdots
(t_{n}^{2}+1)u-1=0$. The constraint ideal is then found as the Gr\"{o}bner
basis of an elimination ideal w.r.t. $u$ (the Rabinowitsch trick \cite%
{Rabinowitsch1929}). This was already formulated as in \cite[Algorithm 1]%
{Li15q}. The constraint ideal (the eliminated ideal) is $I=<f_{1},\ldots
,f_{m}>$ in $\mathbb{R}[\mathbf{t}]$. 

\section{Singularities}

\subsection{Algebraic singularities\label{AS}}

For a single loop linkage or mechanism, by the constraint ideal computation,
we obtain an ideal which we call the constraint ideal $I=<f_{1},\ldots
,f_{m}>$ in the polynomial ring $\mathbb{R}[\mathbf{t}]$ for variables $%
\mathbf{t}$. As mentioned in remark \ref{multiloop}, if a mechanism has
multiple loops, we consider the summation ideal for all loops. Then the
c-space is the solution set $V$ for the polynomial equations where the
polynomials generate the constraint ideal $I$.

When a solution is analyzed algebraically, it is mandatory to assume an
algebraically closed field. Therefore, it is assumed in the following that $%
\mathbb{C}[\mathbf{t}]$ is used. By Hilbert's Nullstellensatz, the vanishing
ideal of $V$ will be the radical ideal $\sqrt{I}$, which can be decomposed
into prime ideals as%
\begin{equation*}
\sqrt{I}=Q_{1}\cap \ldots \cap Q_{s}.
\end{equation*}%
Each of the prime ideals defines a component of $V$ such that%
\begin{equation*}
V=V_{1}\cup \ldots \cup V_{s}.
\end{equation*}%
Also, the ideal $I$ can be decomposed into primary ideals so that%
\begin{equation*}
I=P_{1}\cap \ldots \cap P_{r}.
\end{equation*}%
The primary and prime ideals are not necessarily identical. A simple example
is the following: $I:=<x^{2}>$, $\sqrt{I}=<x>$. This is a very common
phenomenon in modern algebraic geometry for studying a polynomial system,
e.g., schemes defined in algebraic geometry \cite{Grothendieck1964elements,
Shafarevich1994basic} can be employed to deal with such phenomena. In this
paper, we will not talk about schemes, but we are trying to explain this
phenomenon through the procedure of analyzing non-radical ideals which play
an important role in modern algebraic geometry. This should motivate using
modern algebraic tools, e.g. schemes, for kinematic analysis. A simple
example is the following: a primary ideal $I:=<x^{2}>$, and its radical $%
\sqrt{I}=<x>$ is a prime ideal.

A primary decomposition helps in analyzing singularities. In this paper, we
define singularities using the Jacobian matrix of the generators. The \emph{%
algebraic singularities (AS)} are those points where the corank of the
Jacobian matrix form by the generators of the ideal $I$ is bigger than the
local dimension of $V$, and hence they form a closed subset of the $V$. If $%
I $ is a radical ideal, from the decomposition one can distinguish two
principle situations: a point is a singularity if it is a singularity of one
of the $V_{1},\ldots ,V_{s}$ or it is an intersection of the $V_{i}$. Once $%
I $ is not radical, the singularities of the solution variety $V$ do not
reflect all algebraic singularities of the polynomial system, denoted $K$,
defined by the generators of $I$. Those solutions defined by a primary ideal
which is not prime in an irredundant primary decomposition of ideal $I$ will
always be singularities of $K$, where irredundant means: Removing any of
primary ideal changes the intersection; the associated prime ideals $\sqrt{%
P_{i}}$ are distinct. We only take an irredundant primary decomposition in
this paper. A simple example is the following: $I:=<x^{2},xy>=<x^{2},y>\cap
<x>$, $\sqrt{I}=<x>$, where the variety is the $y$-axis. The point $(0,0)$
is an algebraic singularity, which is found by substituting to the Jacobian
matrix defined by two rows $(2x,y)$ and $(0,x)$. The interesting point is
that $I$ is not radical, and the decomposition of the radical $\sqrt{I}$ has
strictly less components than that of $I$. Moreover, the component defined
by $P_{1}=<x^{2},y>$ is embedded in the component defined by $P_{2}=<x>$.

In summary, a first step for checking the algebraic singularities would be
to check whether the constraint ideal $I$ is radical. The second step of the
analysis, for a non-radical constraint ideal $I$, is a primary decomposition
in order to find primary ideals.

Because inhomogenous coordinates are used to formulate the constraints
(Remark~\ref{lt}), even if $I$ is radical, we either use a linear
transformation to get a further ideal that allows checking whether the ideal 
$I$ is radical. One can also achieve this by fixing certain joints.

\subsection{Kinematic Singularities}

A motion of the mechanisms is a real curve satisfying the loop closure
constraints. Therefore, all considerations are made for the real solutions
of $K$.

\begin{defn}
\label{ks} A configuration $\mathbf{t}\in V$ is a kinematic singularity (KS)
iff the rank of the constraint Jacobian $\mathbf{J}$ is not constant in any
real neighborhood in $V$.
\end{defn}

If $\mathbf{t}$ is a kinematic singularity then it is an algebraic
singularity.

\begin{defn}
\label{shaky} A mechanism is shaky iff there is a smooth point of the real
component of $V$ which is not a kinematic singularity and $\dim _{\mathbf{t}%
}V\neq \mathrm{co\mathrm{rank}}~\mathbf{J}\left( \mathbf{t}\right) $, i.e.
it is an AS.
\end{defn}

\begin{rem}
\label{non-radical} If $I$ is not radical, i.e. $\sqrt{I}\neq I$, and if $%
r=s $, and if further a primary ideal $P_{i}$ in the decomposition of $I$
has a real solution, then the mechanism is shaky.
\end{rem}

\section{Example}

In this section, we apply the symbolic framework of analysis of the
singularities to some known examples. All computations are carried out with
the software Maple.

\subsection{Stretched 4-Bar Linkages}

\begin{figure}[t]
\includegraphics[width=0.47\textwidth]{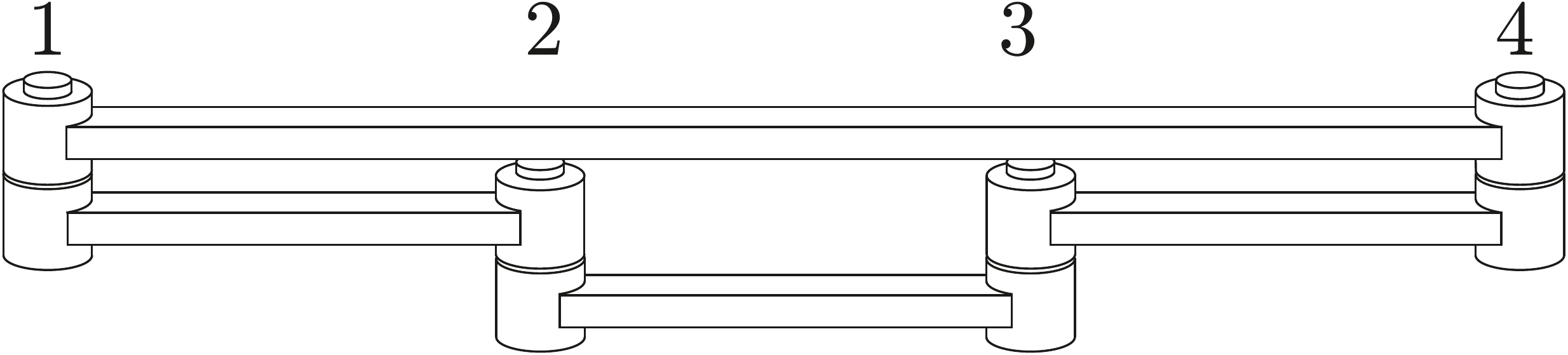}
\caption{Stretched planar 4-bar linkage.}
\label{fig:4r}
\end{figure}

Fig. \ref{fig:4r} shows a planar 4-bar linkage, where the sum of the leghts
of three links equals length of the fourth. This linkage has only one real
configuration and it is shaky. The complex solution space is
one-dimensional. An algebraic explanation for the shakyness is that a point $%
\mathbf{t}$ is always a singularity when the local real dimension of the
configuration set is smaller than the local complex dimension. 

The Denavit-Hartenberg parameters for this example are%
\begin{align*}
(w_{1},w_{2},w_{3},w_{4})& =(0,0,0,0) \\
(d_{1},d_{2},d_{3},d_{4})& =(1,2,3,-6) \\
(s_{1},s_{2},s_{3},s_{4})& =(0,0,0,0).
\end{align*}%
Then the $g_{i}$ in (\ref{gi:1}) for this linkage are%
\begin{equation*}
g_{1}=1-\epsilon \mathbf{k},\ g_{2}=1-2\epsilon \mathbf{k},\
g_{3}=1-3\epsilon \mathbf{k},\ g_{4}=1+6\epsilon \mathbf{k}.
\end{equation*}%
Using the algebraic framework, one obtains the constraint ideal%
\begin{align*}
&
I=<t_{1}t_{2}+3t_{1}t_{3}+2t_{2}t_{3},4t_{1}t_{2}^{2}+8t_{2}^{2}t_{3}+18t_{1}+15t_{2}+9t_{3},
\\
&
-t_{1}t_{2}+6t_{1}t_{4}+5t_{2}t_{4},8t_{2}^{2}t_{3}^{2}+5t_{2}^{2}+6t_{2}t_{3}+9t_{3}^{2},
\\
&
2t_{2}t_{3}+5t_{2}t_{4}+3t_{3}t_{4},8t_{2}^{2}t_{3}+20t_{2}^{2}t_{4}+3t_{2}+9t_{3}+18t_{4}>.
\end{align*}%
This defines a complex 1-dimensional set, but only one real solution. It can
be shown that the ideal $I$ is radical. In fact, by the ideal membership
checking, we have $I=<f_{1},f_{2},f_{3},f_{4}>$, where%
\begin{align*}
f_{1}& =-t_{1}t_{2}+6t_{1}t_{4}+5t_{2}t_{4}, \\
f_{2}& =8t_{2}^{2}t_{3}^{2}+5t_{2}^{2}+6t_{2}t_{3}+9t_{3}^{2}, \\
f_{3}& =4t_{1}t_{2}^{2}+8t_{2}^{2}t_{3}+18t_{1}+15t_{2}+9t_{3}, \\
f_{4}& =8t_{2}^{2}t_{3}+20t_{2}^{2}t_{4}+3t_{2}+9t_{3}+18t_{4}.
\end{align*}%
The Jacobian matrix $\mathbf{J}$ (based on generators $%
f_{1},f_{2},f_{3},f_{4}$ of the ideal) are calculated by differentation
w.r.t. $t_{1},t_{2},t_{3},t_{4}$. Adding the ideal $J$, defined by the $3$%
-minors of $\mathbf{J}$, to the ideal $I$, we find all singularities of
variety of $I$. In fact, it only has one real solution: $(0,0,0,0)$, which
is also the only real configuration for the stretched 4-bar. This shows that
shakiness can happen when constraint ideal is radical, but locally the
dimension of the real solution is smaller than the dimension of the complex
solution, namely the configuration is a smooth point in the real variety but
a singularity in the complex variety of $K$.

\subsection{\label{6r}Special 6R linkages constructed from Goldberg linkages}

\begin{figure}[t]
\includegraphics[width=0.45\textwidth]{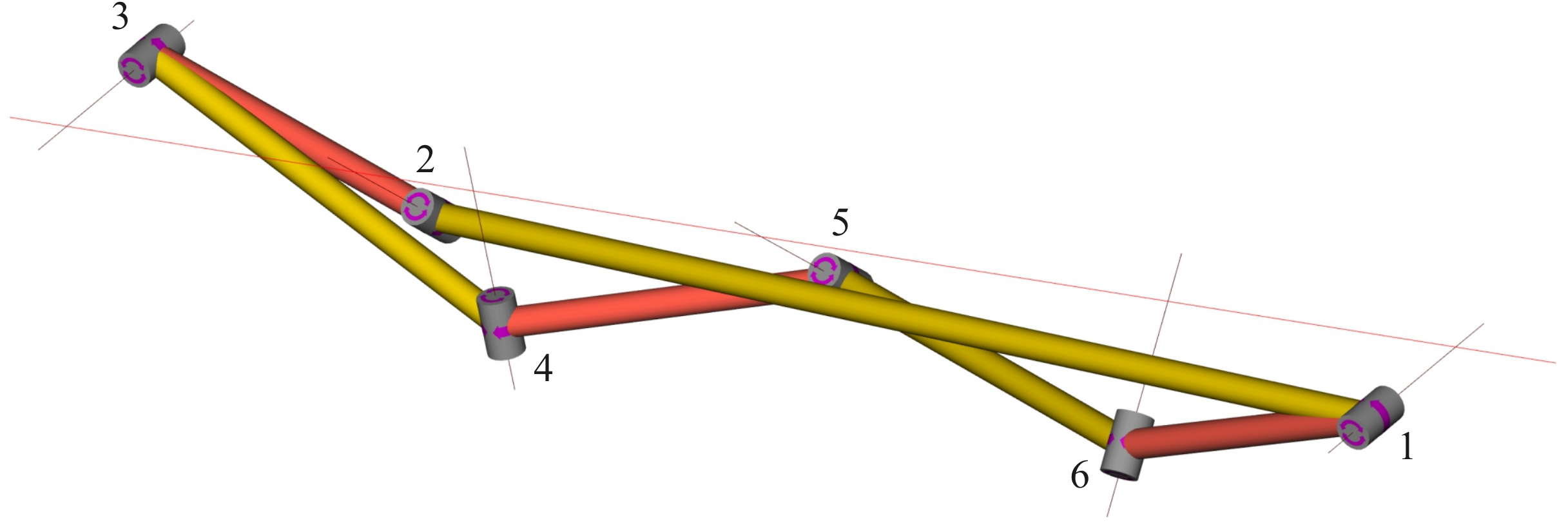}
\caption{A special Goldberg serial 6R linkage.}
\label{fig:6r}
\end{figure}

A special 6R linkage (Fig. \ref{fig:6r}) constructed using Bennett linkages
in \cite{Chen:r12} has special kinematic singularities as discussed in \cite%
{Mueller16local}. The peculiar feature is that its c-space $V$ is a smooth
1-dimensional manifold, but at two points (two configurations) of this curve
the constraint Jacobian drops rank, rendering them as kinematic
singularities. They could be identified by local analysis of the subset of
the c-space of points with a certain rank \cite{Mueller16local}. Yet a
deeper understanding of why these singularities can actually happen is
lacking. The algebraic framework reveals that these singularities appear
because the constraint ideal $I$ is primary. 
We have a typical numerical example from the Denavit-Hartenberg parameters
(as we would prefer rational numbers for fast computation).

The Denavit-Hartenberg parameters of the linkage under consideration are%
\begin{align*}
(w_{1},w_{2},w_{3},w_{4},w_{5},w_{6})& =(1,1,1/3,-1,1/3,1), \\
(d_{1},d_{2},d_{3},d_{4},d_{5},d_{6})& =(1,1,3/5,3,3/5,1), \\
(s_{1},s_{2},s_{3},s_{4},s_{5},s_{6})& =(0,0,0,0,0,0).
\end{align*}%
The corresponding dual quaternions $g_{i}$ for this 6R linkage are 
\begin{align*}
g_{1}& =1+\mathbf{k}(-1/2\epsilon -1)-1/2\epsilon , \\
g_{2}& =1+\mathbf{k}(-1/2\epsilon -1)-1/2\epsilon , \\
g_{3}& =1+\mathbf{k}(-3/10\epsilon -1/3)-1/10\epsilon , \\
g_{4}& =1+\mathbf{k}(-3/2\epsilon +1)+3/2\epsilon , \\
g_{5}& =1+\mathbf{k}(-3/10\epsilon -1/3)-1/10\epsilon , \\
g_{6}& =1+\mathbf{k}(-1/2\epsilon -1)-1/2\epsilon
\end{align*}%
The solution set has only a one-dimensional (real) component: $%
I=<4t_{6}+t_{4},t_{4}t_{5}+8,t_{1}t_{4}^{2}+16t_{1}-6t_{4},t_{2}t_{4}^{2}+4t_{2}+3t_{4},4t_{3}+t_{5},4t_{1}t_{2}-t_{1}t_{4}-2t_{2}t_{4},2t_{1}t_{5}-t_{1}t_{4}+6,t_{2}t_{5}-2t_{2}t_{4}-6>. 
$

Further it turns out that the ideal $I$ is a radical ideal. In fact, by the
ideal membership checking, we have $I=<f_{1},f_{2},f_{3},f_{4},f_{5}>$,
where 
\begin{align*}
f_{1}& =4t_{3}+t_{5},f_{2}=4t_{6}+t_{4},f_{3}=t_{4}t_{5}+8, \\
f_{4}&
=t_{1}t_{4}^{2}+16t_{1}-6t_{4},f_{5}=4t_{1}t_{2}-t_{1}t_{4}-2t_{2}t_{4}.
\end{align*}%
The Jacobian matrix (based on generators $f_{1},f_{2},f_{3},f_{4},f_{5}$ of
the ideal) is computed the differentials w.r.t. $%
t_{1},t_{2},t_{3},t_{4},t_{5},t_{6}$. Adding the ideal $J$ generated by all $%
5$-minors of the Jacobian to the ideal $I$, we find no singularity of
variety of $I$. But the linkage has kinematic singularities \cite%
{Mueller16local}. So there is some information missing in our computation.

If we solve the polynomial equations from the polynomials of $I$, we find
that the solution is%
\begin{equation*}
\{t_{1}=\frac{6t_{4}}{t_{4}^{2}+16},t_{2}=-\frac{3t_{4}}{t_{4}^{2}+4},t_{3}=%
\frac{2}{t_{4}},t_{5}=-\frac{8}{t_{4}},t_{6}=-\frac{t_{4}}{4}\}
\label{inhom}
\end{equation*}%
for some real number of $t_{4}$. By the definition of $t_{i}$, we know that
its value can be $(-\infty ,+\infty )$, where the $\infty $ is also a
configuration which corresponds to the rotation of $180$ degrees. For taking
care of all these configurations where some rotations are 180 degrees, we
might use the homogeneous coordinates to represent the rotation angles.
However, its computation will increase a lot, because we double the number
of variables. There is one way to treat is with less computation, namely, we
make a linear transformation for each rotation parameters $t$ by $t\mapsto
(ct+d)/(at+b)$ with arbitrary real numbers $a,b,c,d$ such that $ad-bc$ is
not zero. For instance, a linear transformation as follows%
\begin{align}
u1& =7/10t_{1}+11/4-\mathbf{i}(1/8t_{1}-8/11),  \notag \\
u2& =2/11t_{2}-9/11-\mathbf{i}(t_{2}+5/3),  \notag \\
u3& =t_{3}+1-\mathbf{i}(-2/9t_{3}+3/10),  \label{linear} \\
u4& =11/4t_{4}+11/4-\mathbf{i}(-3/2t_{4}+5/9),  \notag \\
u5& =2/3t_{5}-2/7-\mathbf{i}(-11/3t_{5}+11/5),  \notag \\
u6& =-3/4t_{6}-11/6-\mathbf{i}(11/7t_{6}+3/2),  \notag
\end{align}%
such that our configuration equation becomes%
\begin{equation*}
u_{1}g_{1}u_{2}g_{2}\cdots u_{6}g_{6}\equiv 1,  %\label{eq:c3}
\end{equation*}%
where the variables are still $t_{1},t_{2},t_{3},t_{4},t_{5},t_{6}$. But the
constraint ideal (replaceing the supplement equation $(t_{1}^{2}+1)\cdots
(t_{6}^{2}+1)u-1$ by $u_{1}\overline{u_{1}}\cdots u_{6}\overline{u_{6}}u-1$,
because the linear transformation) becomes $I^{\prime }$. 
The solution set also has only a one-dimensional component:%
\begin{align*}
\{t_{1}& =\frac{10(307557t_{4}^{2}+1880847t_{4}+1617098)}{%
11(424197t_{4}^{2}+530712t_{4}+113360)},  \notag \\
t_{2}& =-\frac{124983t_{4}^{2}+203013t_{4}+91720}{%
6(9801t_{4}^{2}+18603t_{4}+10171)},  \notag \\
t_{3}& =-\frac{27(357t_{4}+320)}{10(837t_{4}+911)},t_{5}=\frac{%
3(549t_{4}+290)}{35(99t_{4}+62)},  \notag \\
t_{6}& =-\frac{154(189t_{4}+152)}{9(3093t_{4}+2834)}\}  %\label{hom}
\end{align*}%
for some real number of $t_{4}$. 
Further checking for the ideal $I^{\prime }$ reveals that it is not a
radical ideal. In fact, by the ideal membership checking, we have $I^{\prime
}=<f_{1},f_{2},f_{3},f_{4},f_{5},f_{6}>$. 
Adding the ideal $J$ generated by the $5$-minors of the Jacobian matrix
(based on generators $f_{1},f_{2},f_{3},f_{4},f_{5},f_{6}$ of the ideal) to
the ideal $I^{\prime }$, we find all singularities of variety of $I$. The
singularities contain two points%
\begin{align*}
\{t_{1}& =\frac{64}{11},\ t_{2}=-\frac{5}{3},\ t_{3}=\frac{27}{20},\
t_{4}=-1,\ t_{5}=\frac{3}{5},\ t_{6}=-\frac{22}{9}\}, \\
\{t_{1}& =\frac{64}{11},\ t_{2}=-\frac{5}{3},\ t_{3}=-1,\ t_{4}=\frac{10}{27}%
,\ t_{5}=\frac{3}{7},\ t_{6}=-\frac{21}{22}\}.
\end{align*}%
After substituting these two solution to the constraint equations $1-u_{i}%
\mathbf{i}$ with the linear transformation~(\ref{linear}), we have 
\begin{align*}
& [\frac{1501}{220},-\frac{37}{33},\frac{47}{20},-\frac{37}{18}\mathbf{i},%
\frac{4}{35},\frac{295}{126}\mathbf{i}], \\
& [\frac{1501}{220},-\frac{37}{33},-\frac{47}{90}\mathbf{i},\frac{407}{108},-%
\frac{22}{35}\mathbf{i},-\frac{295}{264}].
\end{align*}%
These show that the singularities happen for zero rotations (a real number)
or 180 degree rotations (a scalar times $\mathbf{i}$).
\begin{figure}[b]
\centerline{
\includegraphics[width=0.42\textwidth]{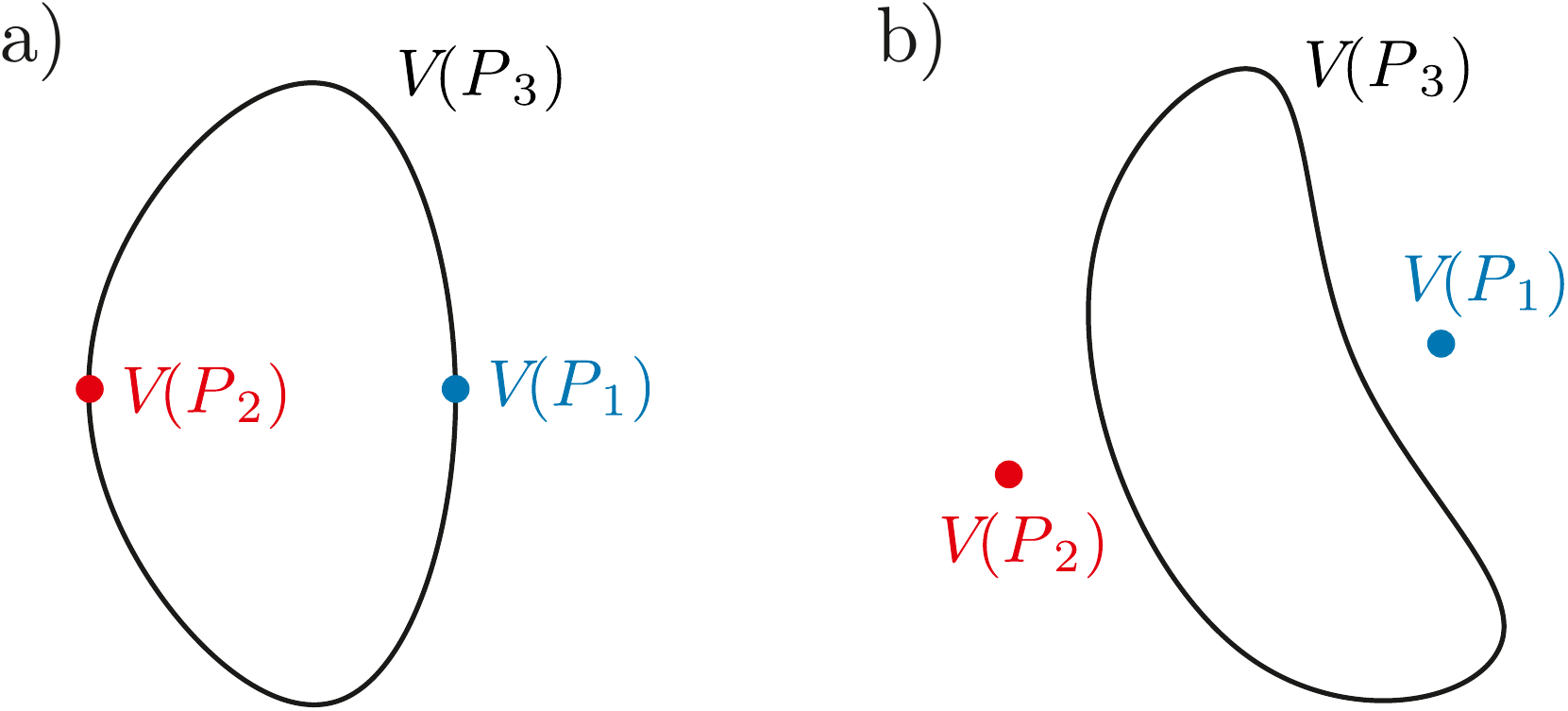}
}
\caption{Schematic representation of the c-space of a) the double Goldberg,
and b) a double Bennett linkage \protect\cite{Selig18double}.}
\label{figGoldberg+Bennett_C-space}
\end{figure}

Further analysis of the non-radical ideal $I^{\prime }$ can be done for
doing a primary decomposition of it, i.e., $I^{\prime }=P_{1}\cap P_{2}\cap
P_{3}$ The decomposition contains three ideals: two non-radical primary
ideals $P_{1}$ and $P_{2}$ (not prime and the varieties are zero-dimensional
and each has just one point) and one prime ideal $P_{3}$ (the 1-dimensional
configuration curve), see Fig. \ref{figGoldberg+Bennett_C-space}a). But the
ideal memberships on their radical ideals shows that $P_{3}\subset \sqrt{%
P_{1}}$ and $P_{3}\subset \sqrt{P_{2}}$. Then the two isolated solutions are
lying on the configuration curve, which means that locally the rank of
Jacobian changed at the solutions.

We see that this 6R linkage is a special of Waldron 6R linkages at \cite%
{Dietmaier95}. Namely, we can find six rotation axes in the space such that
they are the initial configuration of such a 6R linkage. In order to have
polynomials with rational number coefficients, we use another example for
the computation. We have six rotations in terms of dual quaternions as: 
\begin{align*}
h_{1}& =\mathbf{i}, \\
h_{2}& =\mathbf{j}+\frac{\epsilon \mathbf{i}}{2}+\frac{\epsilon \mathbf{k}}{2%
}, \\
h_{3}& =\mathbf{i}\left( \frac{21}{29}-\frac{80}{841}\epsilon \right) -%
\mathbf{j}\left( \frac{20}{29}+\frac{84}{841}\epsilon \right) -\frac{10}{29}%
\epsilon \mathbf{k}, \\
h_{4}& =\mathbf{i}\left( -\frac{144}{145}+\frac{2159}{42050}\epsilon \right)
+\mathbf{j}\left( \frac{17}{145}+\frac{9144}{21025}\epsilon \right) +\frac{17%
}{290}\epsilon \mathbf{k}, \\
h_{5}& =\mathbf{i}\left( -\frac{13}{85}-\frac{1008}{7225}\epsilon \right) +%
\mathbf{j}\left( \frac{84}{85}-\frac{156}{7225}\epsilon \right) +\frac{42}{85%
}\epsilon \mathbf{k}, \\
h_{6}& =\mathbf{i}\left( -\frac{8}{17}+\frac{225}{578}\epsilon \right) -%
\mathbf{j}\left( \frac{15}{17}+\frac{60}{289}\epsilon \right) -\frac{15}{34}%
\epsilon \mathbf{k}.
\end{align*}%
By the symbolic framework, 
using an arbitrary linear transformation as before, we can find the two
singularities and primary ideals. It is worth to point out that this is a
special case 
of the double Bennett intersection as discussed in \cite{Selig18double}, see
Fig. \ref{figGoldberg+Bennett_C-space}b). The solution set of the latter
consists of a conic and two isolated points. In this 6R linkage, the two
isolated points lie on the conic. The degree counting here is made in the
Study quadric. One can quickly check that the Jacobian matrix defined by the
Pl\"{u}cker coordinates \cite{Karger1996singularity,Li18ark} has rank four at these two singular
configurations. It is known from the line geometry \cite%
{Merlet1992geometry,Nawratil04geometrie,Pottmann09line} this rank will become three if all axes
lie on a plane. The 6R linkages constructed by the third type of Bricard
octahedra \cite{Bricard1897,Bricard1926leccons,Baker1980bricard,Baker2009skew} have two
co-planar configurations (all joints are lying in one plane), where the
configurations are also defined by two more primary ideals (not prime) in an
irredundant primary decomposition of the constraint ideal which are revealed
by our symbolic framework. Examples analysis for demonstration is shown in
the appendix.

These examples show that a mechanism can have kinematic singularities even
when the variety of $K$ is a (complex as well as real) manifold. The
singularity is not revealed by the differential geometry. This can only be
revealed by checking whether constraint ideal $I$ is radical. Furthermore,
the radical ideal of a primary component of a decomposition of the
constraint ideal $I$ contains another prime ideal of that decomposition,
i.e., $\sqrt{P_{i}}\supset P_{j}$. A geometrical explanation is that a
double point defined by a primary ideal is embedded in a smooth curve.

\subsection{A special shaky 7R linkage}

\begin{figure}[!t]
\includegraphics[width=0.47\textwidth]{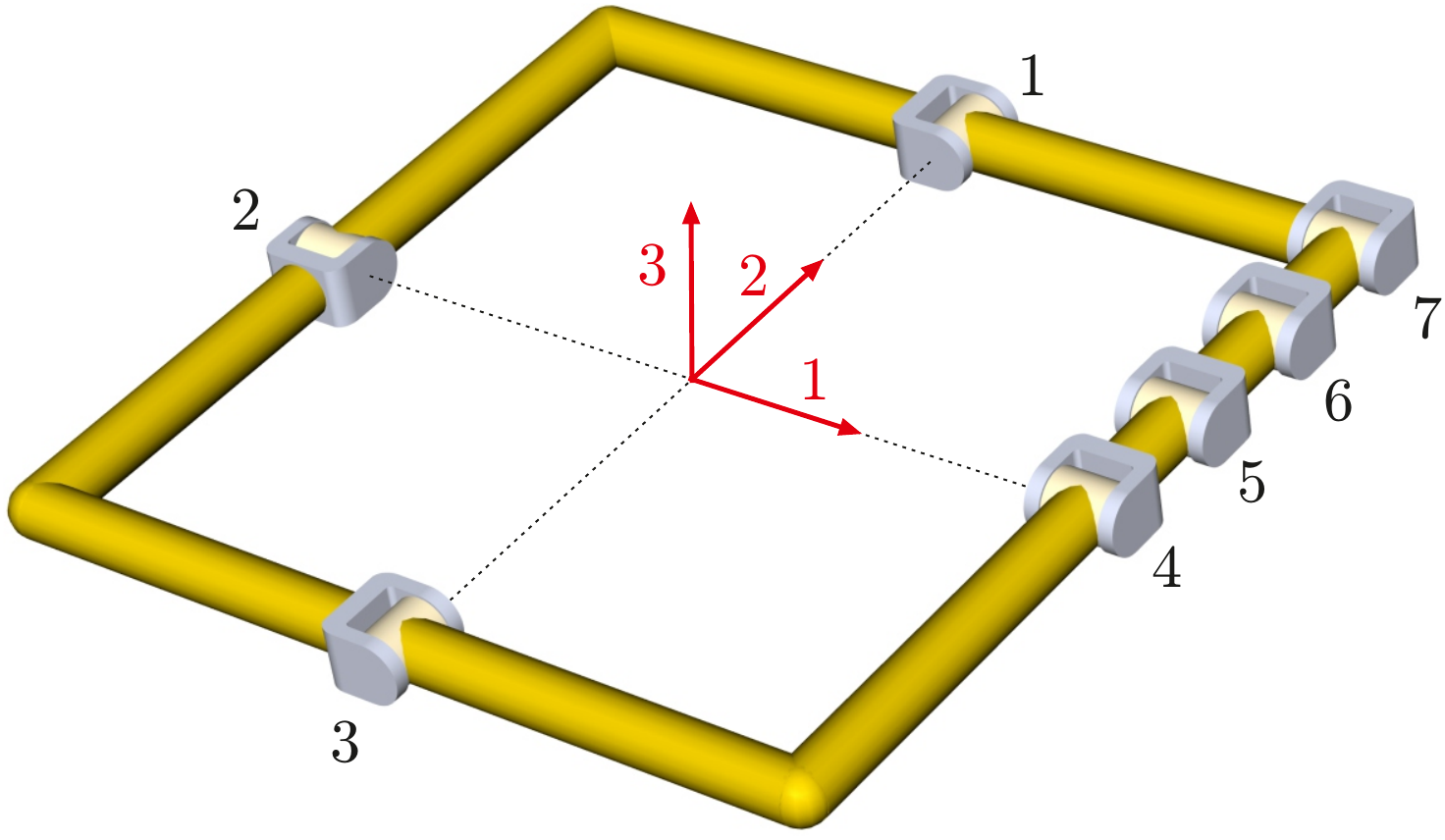}
\caption{A shaky 7R linkage.}
\label{fig:7r}
\end{figure}

A special 7R linkage (Fig.~\ref{fig:7r}) constructed using planar straight
4bar as a part of the chain has special kinematic singularities as discussed
in \cite{Mueller16local}. 

The Denavit-Hartenberg parameters can be deduced from Fig.~\ref{fig:7r}. The
dual quaternions for the seven rotation axes in the initial configuration
are (one can change the world coordinates to get other dual quaternions): 
\begin{align*}
% \begin{eqnarray}               
h_{1}& =\mathbf{i},\quad h_{2}=\mathbf{j},\quad h_{3}=\mathbf{i},\quad h_{4}=%
\mathbf{j}-3\epsilon \mathbf{k}, \\
h_{5}& =\mathbf{j}-2\epsilon \mathbf{k},\quad h_{6}=\mathbf{j}-\epsilon 
\mathbf{k},\quad h_{7}=\mathbf{j}.
\end{align*}%
%               \end{eqnarray}
We have a constraint ideal: % \begin{equation*}
$%
I_{1}=<15513t_{4}-3864t_{3}-8410t_{6},5171t_{5}-2023t_{3}-6745t_{6},2t_{3}t_{6}-t_{6}^{2},4t_{3}^{2}-t_{6}^{2},2t_{1}t_{3}-t_{1}t_{6},2t_{2}t_{3}-t_{2}t_{6},t_{1}t_{2}+3t_{1}t_{6}+8t_{2}t_{6},15513t_{1}t_{6}^{2}+20684t_{1}+8624t_{3}-24996t_{6},10342t_{2}t_{6}^{2}+15513t_{2}-4248t_{3}+7295t_{6}>. 
$ % \end{equation*}

The ideal $I$ is found to be radical. Using another arbitrary linear
transformation as before, we find that its constraint ideal is also radical.
It is similar to the stretched 4bar whose singularities are those
configurations where the local real dimension is smaller than the local
complex dimension.

\subsection{A special shaky 4R-5R linkage}

\begin{figure}[t]
\includegraphics[width=0.47%
\textwidth]{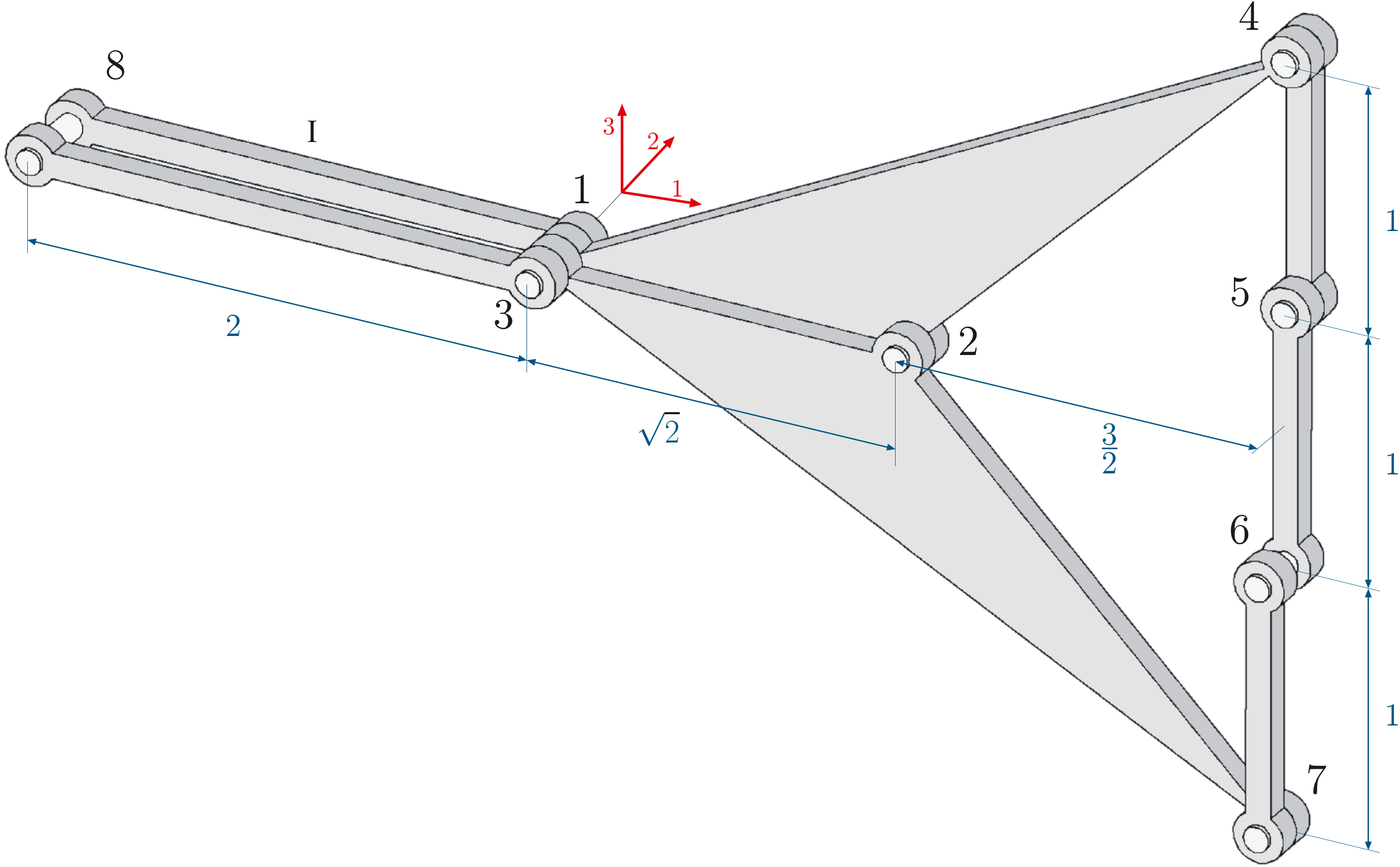}
\caption{A shaky extended Watt linkage.}
\label{fig:45r}
\end{figure}

A special 4R-5R linkage (extended Watt linkage) in Fig.~\ref{fig:45r}
constructed using planar a stretched 4-bar as a part of the chain has
special kinematic singularities as discussed in \cite{Mueller16local}. It is
a kinematotropic linkage, wherein its 1-dimensional motion mode the 4-bar
sub-linkage remains stretched, thus shaky. 
The symbolic framework for analyzing the kinematic singularities is applied
to analyze the dimension drop at some real configuration (singularities).

The Denavit-Hartenberg parameters are indicated in Fig.~\ref{fig:7r}. We can
even write a sequence of rotation axes using dual quaternions for
representing the initial configuration from the figure. The eight axes (with
a rational number preferences) can be (one joint is shared by two loops): 
\begin{align*}
h_{1}& =\mathbf{i},h_{2}=\mathbf{i}+2\epsilon \mathbf{k},h_{3}=\mathbf{i}%
,h_{4}=\mathbf{i}-\frac{6}{7}\epsilon \mathbf{k}, \\
h_{5}& =\mathbf{i}-\left( \frac{6}{7}+\frac{3}{2}\right) \epsilon \mathbf{k}-%
\frac{3}{2}\epsilon \mathbf{j},h_{6}=\mathbf{i}-\left( \frac{6}{7}+\frac{3}{2%
}\right) \epsilon \mathbf{k}-\frac{1}{2}\epsilon \mathbf{j}, \\
h_{7}& =\mathbf{i}-\left( \frac{6}{7}+\frac{3}{2}\right) \epsilon \mathbf{k}+%
\frac{1}{2}\epsilon \mathbf{j},h_{8}=\mathbf{i}-\left( \frac{6}{7}+\frac{3}{2%
}\right) \epsilon \mathbf{k}+\frac{3}{2}\epsilon \mathbf{j}.
\end{align*}

By the symbolic framework with $u_{1}u_{2}\cdots u_{n}\equiv 1$, where $%
u_{i} $ is obtained by a linear transformation of $t_{s}$ for $h_{s}$ for $%
s=1,\ldots ,n$, we have two constraint ideals $I_{l}$ for the left loop and $%
I_{r}$ for the right loop. The constraint ideal $I$ for the mechanism will
be the summation $I:=I_{l}+I_{r}$. Further checking for the ideal $I$
reveals that it is a radical ideal. With further Jacobian computation as
before for the generators of $I$, the singularities are also similar to the
stretched 4bar whose singularities are those real configurations with lower
dimension locally compare to its complex dimension.

\subsection{A special shaky planar linkage}

\begin{figure}[t]
\includegraphics[width=0.4\textwidth]{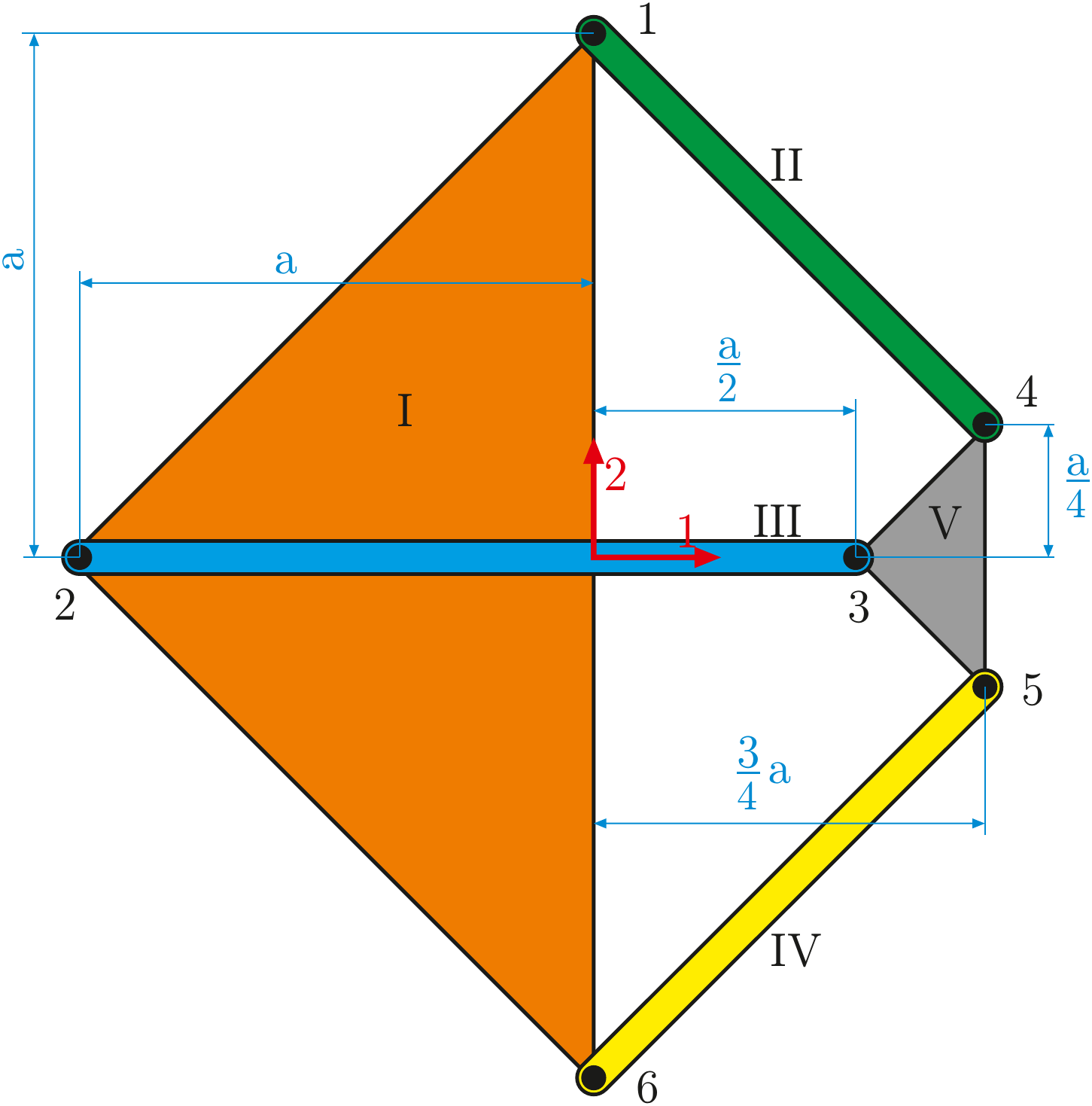}
\caption{A shaky linkage reported by Wohlhart \protect\cite%
{Wohlhart99degrees}}
\label{fig:44r}
\end{figure}

A special planar shaky linkage was presented by \cite{Wohlhart99degrees}
(Fig. \ref{fig:44r}) is discussed in \cite{Mueller16local}.

The Denavit-Hartenberg parameters are deduce from Fig.~\ref{fig:44r}. The
dual quternions for the rotation axes in the initial configuration are%
\begin{align*}
h_{1}& =\mathbf{i}+a\epsilon \mathbf{j}+a\epsilon \mathbf{k},\quad h_{2}=%
\mathbf{i},\quad h_{3}=\mathbf{i}+\frac{3a}{2}\epsilon \mathbf{j}, \\
h_{4}& =\mathbf{i}+\frac{7a}{4}\epsilon \mathbf{j}+\frac{a}{4}\epsilon 
\mathbf{k},h_{5}=\mathbf{i}+\frac{7a}{4}\epsilon \mathbf{j}-\frac{a}{4}%
\epsilon \mathbf{k},h_{6}=\mathbf{i}+a\epsilon \mathbf{j}-a\epsilon \mathbf{k%
},
\end{align*}%
for a positive real scalar $a$.

By the symbolic framework with $u_{1}u_{2}\cdots u_{n}\equiv 1$, where $%
u_{i} $ is a linear transformation of $t_{i}$, we have two constraint ideals 
$I_{u} $ for the up loop and $I_{r}$ for the down loop. The constraint ideal 
$I$ for the mechanism will be the summation $I:=I_{l}+I_{r}$. Further
checking for the ideal $I$ reveals that it is not a radical ideal. With
further Jacobian computation as before for the generators of $I$, the
singularities are exactly located at a real isolated point (the only real
configuration). It is not counted as a kinematic singularity because locally
the rank is the same.

\section{Conclusion}

Kinematic singularities of mechanisms can appear at a configuration which is
a smooth point of the configuration curve. Using an algebraic framework, an
explanation is given, namely that the constraint ideal which is generated by
the constraint equations is not a radical ideal. With a primary
decomposition, the number of primary ideals in its decomposition is strictly
bigger than the number of varieties in an irreducible decomposition of the
variety of the constraint ideal (the c-space). This approach is applied to
several examples. The special phenomenon only appears in one example, a
special 6R linkage. It is still open to classify all mechanisms with this
special phenomenon, even within mobile 6R linkage.

\section{Acknowledgement}

The authors would like to thank Josef Schicho for discussion and helpful
remarks on algebraic geometry aspect. The authors would like to thank Georg
Nawratil for discussion and helpful remarks on line geometry aspect. The
first author was supported by the Austrian Science Fund (FWF) : P 31061 (The
Algebra of Motions in 3-Space). The first author was
 partially funded by the Austrian Science Fund (FWF): W1214-N15, project DK9.
 The second author acknowledges that this
work has been supported by the LCM-K2 Center within the framework of the
Austrian COMET-K2 program. The authors acknowledge that this work is an
outcome of the Rigidity and Flexibility of Geometric Structures workshop at
the Erwin Schr\"{o}dinger International Institute for Mathematics and
Physics.

\bibliographystyle{plain}
\bibliography{singular}

\appendix

\section{Appendix}

In this appendix, we will give a general method to find special mobile 6R
linkage with the typical phenomenon which is discussed in Section~\ref{6r}
using some well-known results from line geometry.

In line geometry, the $6\times 6$ matrix from Pl\"{u}cker coordinates of six
lines which lie on one plane has rank 3. For a mobile 6R linkage, a
configuration when six rotation axes are lying on one plane will be
singular. In other words, the configuration space is singular. As discussed
in this paper, we are interested in which situation of a singularity can be.
The symbolic framework investigates several 6R linkages with this particular
singularity. The computation reveals that the phenomenon appeared in Section~%
\ref{6r} is not single. However, we still have to mention that we only find
some surficient conditions for having a non-radical constraint ideal. The
first reason is that we could not go through all known mobile 6R linkage
which is either not known. The second reason is that the rank of the
Jacobian matrix at the particular singularities in Section~\ref{6r} is 4.

A 6R linkage from the type III Bricard octahedron always has two
configurations where all axes lie on a plane. For demonstration, we take such a 6R
linkage with six rotations in terms of dual quaternions as: 
%%%%%%%%%%%%%%%%%%%%%%%%%%%%%%%%%%%%%%%%
\begin{eqnarray}  \label{cf0}
h_{1}& =&\frac{5}{13}\mathbf{i} - \frac{12}{13}\mathbf{j}, \quad h_{2} =\frac{63}{65}\mathbf{i} -\frac{63}{65}\mathbf{j}, \notag \\
h_{3}& =&\frac{84}{85}\mathbf{i} -\frac{13}{85}\mathbf{j} - \frac{25}{51}\epsilon
\mathbf{k}, \quad
h_{4} = \frac{15}{17}\mathbf{i} + \frac{8}{17}\mathbf{j}
- \frac{4}{17}\epsilon\mathbf{k}, \\
h_{5}& =&\frac{4}{5}\mathbf{i} + \frac{3}{5}\mathbf{j} + \frac{1}{5}\epsilon%
\mathbf{k}, \quad 
h_{6} =\frac{3}{5}\mathbf{i} - \frac{4}{5}\mathbf{j} + \frac{%
4}{15}\epsilon\mathbf{k}. \notag
\end{eqnarray}
Using an arbitrary linear transformation as before, we can find two
singularities and two primary ideals from its non-radical constraint ideal.
These two singular configurations are just the folding configurations where
all rotation axes are on a plane. The first configuration is just the
starting configuration of six rotations in terms of dual quaternions as in %
(\ref{cf0}). The second configuration is just when the rotation angles go to
180 degrees. If we fix the link between the first two joints $h_1$ and $h_2$%
, after rotating one joint with 180 degrees, we will have the rest four
rotation axes on the same plane as: 
\begin{eqnarray*}  
h^{\prime }_{3} & =& \frac{13524}{14365}\mathbf{i} -\frac{4843}{14365}\mathbf{j}
+ \frac{25}{51}\epsilon\mathbf{k}, \
h^{\prime }_{4} = \frac{2745}{2873}\mathbf{i} + \frac{848}{2873}\mathbf{j}
+ \frac{28}{51}\epsilon\mathbf{k}, \notag \\
h^{\prime }_{5} & =& \frac{123}{845}\mathbf{i} - \frac{836}{845}\mathbf{j} + 
\frac{4}{15}\epsilon\mathbf{k}, \
h^{\prime }_{6} =\frac{836}{845}\mathbf{i} + \frac{123}{845}\mathbf{j} + 
\frac{1}{15}\epsilon\mathbf{k}. \notag
\end{eqnarray*}
One can quickly check that the Jacobian matrix defined by the Pl\"ucker
coordinates has rank three at these two configurations. Besides, this type
of 6R linkage is a particular of angle-symmetric 6R linkages (third type) at 
\cite{Li13angle} because it also can follow a cubic motion and it is also
angle-symmetric, but with particular Denavit-Hartenberg parameters, e.g.,
twist distances are zeros. It is worth to mention that the general equations
defining the third type of angle-symmetric 6R linkage are still unknown.

A 6R linkage from the type I Bricard octahedron, in general, does not have a
configuration where all axes lie on a plane. We know that it is a special
Bricard line symmetric 6R linkage. A general Bricard line symmetric 6R
linkage can be constructed as: 1) Take three rotations in terms of dual
quaternions $h_1,h_2,h_3$, where $h_i^2=-1$ for $i = 1,2,3$ . 2) Take a
rotation $u$ with $u^2=-1$ whose rotation axes is perpendicular the plane $%
\mathbf{P}$. 3) Calculate other three rotation by $h4 := -uh_1u$, $h5 :=
-uh_2u$, $h6 := -uh_3u$. 4) Then $L:=[h_1,h_2,h_3,h_4,h_5,h_6]$ is a Bricard
line symmetric 6R linkage. Here if we constraint the first three rotation
such that their axes are lying on a plane $\mathbf{P}$ and the axis of $u$
is perpendicular to the plane $\mathbf{P}$, the linkage $L$ will have a
singular configuration which is exactly the starting configuration. A
numerical computation reveals that the constraint ideal is also non-radical.
One can also check for the 6R linkages from type II Bricard octahedra, the
6R linkage with translation property in \cite{Li13ck}. With our computation,
they all have similar phenomena as in Section~\ref{6r} when we constrain
them to have a configuration with all axes lie on a plane. The particular
singular configuration appears only once which is the starting
configuration. Therefore, we can find many 6R linkages with the particular
phenomenon appears in Section~\ref{6r} if we check for more known 6R
linkages. 

\vspace{2cm}
% \newpage
	{\bf Zijia Li}
	{Research Intitute for Symbolic Computation\\
	Johannes Kepler University, Linz, Austria\\
	\url{zijia.li@risc.jku.at}}\\

	{\bf Andreas M\"uller}
	{Institute of Robotics\\
	Johannes Kepler University, Linz, Austria\\
	\url{a.mueller@jku.at}}

\end{document}